\documentclass{article}
\usepackage{times}
\usepackage{graphicx} 
\usepackage{natbib}
\usepackage[vlined,ruled]{algorithm2e}
\usepackage{bm}
\usepackage{multirow}
\usepackage{amssymb}
\usepackage{cleveref}
\usepackage{epsfig}
\usepackage{times,amsmath,float,amsthm} 
\usepackage{epsf}
\usepackage{xcolor}
\usepackage[lofdepth,lotdepth]{subfig}

\usepackage[accepted, nohyperref]{icml2017}

\def\l0{\ell^0(\mathbb{Z}^{2})}

\def\R{\mathbb R}

\def\RR{\rm \hbox{I\kern-.2em\hbox{R}}}
\def\NN{\rm \hbox{I\kern-.2em\hbox{N}}}
\def\ZZ{\rm {{\rm Z}\kern-.28em{\rm Z}}}

\def\<{\langle}
\def\>{\rangle}

\def\Chi{\raise .3ex \hbox{\large $\chi$}}

\def\v1{\tilde v}

\def\[{\Bigl [}
\def\]{\Bigr ]}
\def\({\Bigl (}
\def\){\Bigr )}
\def\[{\Bigl [}
\def\]{\Bigr ]}
\def\({\Bigl (}
\def\){\Bigr )}

\newcommand{\bz}{\mathbf{z}}
\newcommand{\bw}{\mathbf{w}}
\newcommand{\be}{\begin{equation}}
\newcommand{\ee}{\end{equation}}
\newcommand{\beq}{\begin{eqnarray}}
\newcommand{\eeq}{\end{eqnarray}}
\newcommand{\bea}{\begin{array}{llll}}
\newcommand{\eea}{\end{array}}
\newcommand{\bi}{\begin{itemize}}
\newcommand{\ei}{\end{itemize}}
\newcommand{\ie}{i.e.}
\newcommand{\MovieL}{\textsc{MovieLens}}

\newcommand{\argmin}{\rm argmin}

\newcommand{\argmax}{\rm argmax}

\newcommand{\CCOT}{\texttt{CCOT}}
\newcommand{\CCOTGW}{\texttt{CCOT-GW}}

\icmltitlerunning{Co-clustering through Optimal Transport}

\begin{document} 

\twocolumn[
\icmltitle{Co-clustering through Optimal Transport}


\begin{icmlauthorlist}
\icmlauthor{Charlotte Laclau}{ed}
\icmlauthor{Ievgen Redko}{creatis}
\icmlauthor{Basarab Matei}{ed}
\icmlauthor{Youn\`es Bennani}{ed}
\icmlauthor{Vincent Brault}{greb}
\end{icmlauthorlist}

\icmlaffiliation{creatis}{CNRS UMR 5220 - INSERM U1206, Univ. Lyon 1, INSA Lyon, F-69621 Villeurbanne, France }
\icmlaffiliation{greb}{CNRS, LJK, Univ. Grenoble-Alpes, France}
\icmlaffiliation{ed}{CNRS, LIPN, Universit\'e Paris 13 - Sorbonne Paris Cit\'e, France}

\icmlcorrespondingauthor{Charlotte Laclau}{charlotte.laclauc@univ-grenoble-alpes.fr}

\vskip 0.3in
]
\printAffiliationsAndNotice{} 
\footnotetext{The first author of this paper is now a post-doc in CNRS, LIG, Univ. Grenoble-Alpes, France }

\begin{abstract} 
In this paper, we present a novel method for co-clustering, an unsupervised learning approach that aims at discovering homogeneous groups of data instances and features by grouping them simultaneously. 
The proposed method uses the entropy regularized optimal transport between empirical measures defined on data instances and features in order to obtain an estimated joint probability density function represented by the optimal coupling matrix. This matrix is further factorized to obtain the induced row and columns partitions using multiscale representations approach. To justify our method theoretically, we show how the solution of the regularized optimal transport can be seen from the variational inference perspective thus motivating its use for co-clustering. The algorithm derived for the proposed method and its kernelized version based on the notion of Gromov-Wasserstein distance are fast, accurate and can determine automatically the number of both row and column clusters. These features are vividly demonstrated through extensive experimental evaluations. 
\end{abstract} 

\section{Introduction}

Cluster analysis aims to gather data instances into groups, called clusters, where instances within one group are similar among themselves while instances in different groups are as dissimilar as possible. Clustering methods have become more and more popular recently due to their ability to provide new insights into unlabeled data that may be difficult or even impossible to capture for a human being. Clustering methods, however, do not take into account the possible existing relationships between the features that describe the data instances. For example, one may consider a data matrix extracted from text corpus where each document is described by the terms appearing in it. In this case, clustering documents may benefit from the knowledge about the correlation that exists between different terms revealing their probability of appearing in the same documents. This idea is the cornerstone of \textit{co-clustering} \cite{hartigan-direct-clustering-data-1972,opac-b1078049} where the goal is to perform clustering of both data points and features simultaneously. The obtained latent structure of data is composed of blocks usually called co-clusters. Applications of co-clustering include but are not limited to recommendation systems \cite{merugu_05,deodhar10,Xu:2012:EIC:2187836.2187840}, gene expression analysis \cite{cheng_08,citeulike:350992} and text mining \cite{dhillon_03,pu2009latent}. As a result, these methods are of an increasing interest to the data mining community.  

Co-clustering methods are often distinguished into probabilistic methods (e.g., \cite{Dhillon:2003:IC:956750.956764, Banerjee:2007:GME:1314498.1314563, NADI08CI, pu2009latent, shan_10}) and metric based (e.g., \cite{rocci_08, ding_06}) methods
. Probabilistic methods usually make an assumption that data was generated as a mixture of probability density functions where each one of them corresponds to one co-cluster. The goal then is to estimate the parameters of the underlying distributions and the posterior probabilities of each co-cluster given the data. Metric based approaches proceed in a different way and rely on introducing and optimizing a criterion commonly taking into account intra- and inter-block variances. This criterion, in its turn, is defined using some proper metric function that describes the geometry of data in the most precise way possible. Both metric and probabilistic approaches are known to have their own advantages and limitations: despite being quite efficient in modeling the data distribution, probabilistic methods are computationally demanding and hardly scalable; metric methods are less computationally demanding but present the need to choose the ``right" distance that uncovers the underlying latent co-clusters' structure based on available data. Furthermore, the vast majority of co-clustering methods require the number of co-clusters to be set in advance. This is usually done using the computationally expensive exhaustive search over a large number of possible pairs of row and column clusters as in \cite{KeribinBCG15,WyseF12a,wyse14}.

In this paper, we address the existing issues of co-clustering methods described above by proposing a principally new approach that efficiently solves the co-clustering problem from both qualitative and computational points of view and allows the automatic detection of the number of co-clusters. 
We pose the co-clustering problem as the task of transporting the empirical measure defined on the data instances to the empirical measure defined on the data features. The intuition behind this process is very natural to co-clustering and consists in capturing the associations between instances and features of the data matrix. The solution of optimal transportation problem is given by a doubly-stochastic coupling matrix which can be considered as the approximated joint probability distribution of the original data. Furthermore, the coupling matrix can be factorized into three terms where one of them reflects the posterior distribution of data given co-clusters while two others represent the approximated distributions of data instances and features. We use these approximated distributions to obtain the final partitions. We also derive a kernelized version of our method that contrary to the original case, is based on an optimal transportation metric defined on the space of dissimilarity functions. 

The main novelty of our work is two-fold. To the best of our knowledge, the proposed approach is a first attempt to apply entropy regularized optimal transport for co-clustering and to give its solution a co-clustering interpretation. While Wasserstein distance has already been adapted to design clustering algorithms \cite{DBLP:conf/icml/CuturiD14,Irpino:2014}, our idea is to concentrate our attention on the solution of the optimal transport given by the coupling matrix and not to minimize the quantization error with respect to (w.r.t.) Wasserstein distance. We also note that using entropy regularization leads to a very efficient algorithm that can be easily parallelized \cite{conf/nips/Cuturi13}. Second, we show that under some plausible assumptions the density estimation procedure appearing from the use of the optimal transport results in the variational inference problem with the minimization of the reversed Kullback-Leibler divergence. The important implications of this difference w.r.t. other existing methods are explained in Section 3. 

The rest of this paper is organized as follows. In Section 2, we briefly present the discrete version of the optimal transportation problem and its entropy regularized version. Section 3 proceeds with the description of the proposed approach, its theoretical analysis and algorithmic implementation. In Section 4, we evaluate our approach on synthetic and real-world data sets and show that it is accurate and substantially more efficient than the other state-of-the-art methods. Last section concludes the paper and gives a couple of hints for possible future research. 

\section{Background and notations}
In this section, we present the formalization of the Monge-Kantorovich \cite{kantorovich} optimization problem and its entropy regularized version.

\subsection{Optimal transport} 
Optimal transportation theory was first introduced in \cite{monge_81} to study the problem of resource allocation. Assuming that we have a set of factories and a set of mines, the goal of optimal transportation is to move the ore from mines to factories in an optimal way, \ie , by minimizing the overall transport cost. 

More formally, given two empirical probability measures\footnote{Due space limitation, we present only the discrete version of optimal transport. For more details on the general continuous case and the convergence of empirical measures, we refer the interested reader to the excellent monograph by \cite{opac-b1129524}.} $\hat{\mu}_S = \frac{1}{N_S}\sum_{i=1}^{N_S}\delta_{x_i^S}$ 
and $\hat{\mu}_T = \frac{1}{N_T}\sum_{i=1}^{N_T}\delta_{x_i^T}$ 
defined as uniformly weighted sums of Dirac with mass at locations supported on two point sets $X_S = \{\bm{x}^{S}_i \in \mathbb{R}^d\}_{i=1}^{N_S}$ and $X_T = \{\bm{x}^{T}_i \in \mathbb{R}^d\}_{i=1}^{N_T}$, the Monge-Kantorovich problem consists in finding a probabilistic coupling $\gamma$ defined as a joint probability measure over 
$X_S \times X_T$
with marginals $\hat{\mu}_S$ and $\hat{\mu}_T$ that minimizes the cost of transport w.r.t. some metric $l:X_s \times X_t \rightarrow \mathbb{R}^+$: 
$$\min_{\gamma \in \Pi(\hat{\mu}_S, \hat{\mu}_T)}\langle M, \gamma\rangle_F$$
where $\langle \cdot \text{,} \cdot \rangle_F$ is the Frobenius dot product, $\Pi(\hat{\mu}_S, \hat{\mu}_T) = \lbrace \gamma \in \mathbb{R}^{N_S \times N_T}_+ \vert \gamma \bm{1} = \hat{\mu}_S, \gamma^T \bm{1} = \hat{\mu}_T\rbrace$ is a set of doubly stochastic matrices and $M$ is a dissimilarity matrix, \ie, $M_{ij} = l(\bm{x}_i^S,\bm{x}_j^T)$, defining the energy needed to move a probability mass from $\bm{x}_i^S$ to $\bm{x}_j^T$. This problem admits a unique solution $\gamma^*$ and defines a metric on the space of probability measures (called the Wasserstein distance) as follows:
$$W(\hat{\mu}_S, \hat{\mu}_T) = \min_{\gamma \in \Pi(\hat{\mu}_S, \hat{\mu}_T)}\langle M, \gamma\rangle_F.$$
The Wasserstein distance has been successfully used in various applications, for instance: computer vision \cite{Rubner:2000:EMD:365875.365881}, texture analysis \cite{conf/scalespace/RabinPDB11}, tomographic reconstruction \cite{abraham_2015}, domain adaptation \cite{courty:hal-01018698}, metric learning \cite{Cuturi:2014:GML:2627435.2627452} 
and clustering \cite{DBLP:conf/icml/CuturiD14,Irpino:2014}. This latter application is of a particular interest as Wasserstein distance is known to be a very efficient metric due to its capability of taking into account the geometry of data through the pairwise distances between samples.
The success of algorithms based on this distance is also due to \cite{conf/nips/Cuturi13} who introduced an entropy regularized version of optimal transport that can be optimized efficiently using matrix scaling algorithm. We present this regularization below. 

\subsection{Entropic regularization}
The idea of using entropic regularization dates back to \cite{schrodinger}. In \cite{conf/nips/Cuturi13}, it found its application to the optimal transportation problem through the following objective function:
\begin{align*}
    \min_{\gamma \in \Pi(\hat{\mu}_S,\hat{\mu}_T)}\langle M, \gamma\rangle_F - \frac{1}{\lambda}E(\gamma).
\end{align*}
Second term $E(\gamma) = -\sum_{i,j}^{N_S,N_T} \gamma_{i,j}\log(\gamma_{i,j})$ in this equation allows to obtain smoother and more numerically stable solutions compared to the original case and converges to it at the exponential rate \cite{2015-Benamou-Bregman}. Another advantage of entropic regularization is that it allows to solve optimal transportation problem efficiently using Sinkhorn-Knopp matrix scaling algorithm \cite{sinknopp_67}. 

In the next section, we explain the main underlying idea of our approach that consists in associating data instances with features through regularized optimal transport.

\section{Co-clustering through optimal transport}\label{sec:ccw}
In this section we show how the co-clustering problem can be casted in a principally new way and then solved using the ideas from the optimal transportation theory.

\subsection{Problem setup}
Let us denote by $X$ and $Y$ two random variables taking values in the sets $\{\bm{x}_r\}_{r=1}^{n}$ and $\{\bm{y}_c\}_{c=1}^{d}$, respectively, where subscripts $r$ and $c$ correspond to rows (instances) and columns (features). Similar to \cite{Dhillon:2003:IC:956750.956764}, we assume that the joint probability distribution between $X$ and $Y$ denoted by $p(X,Y)$ is estimated from the data matrix $\mathcal{A} \in \mathbb{R}^{n \times d}$. We further assume that $X$ and $Y$ consist of instances that are distributed w.r.t. probability measures $\mu_r$, $\mu_c$ supported on $\Omega_r, \Omega_c$ where $\Omega_r \subseteq \mathbb{R}^d$ and $\Omega_r \subseteq \mathbb{R}^n$, respectively. 

The problem of co-clustering consists in jointly grouping the set of features and the set of instances into homogeneous blocks by finding two assignment functions $C_r$ and $C_c$ that map as follows: $C_r: \{\bm{x}_1, \dots, \bm{x}_n \} \rightarrow \{\hat{x}_1, \dots, \hat{x}_g \}$, $C_c: \{\bm{y}_1, \dots, \bm{y}_d \} \rightarrow \{\hat{y}_1, \dots, \hat{y}_m \}$ where $g$ and $m$ denote the number of row and columns clusters, and discrete random variables $\hat{X}$ and $\hat{Y}$ represent the partitions induced by $X$ and $Y$, \ie, $\hat{X} = C_r(X) = $ and $\hat{Y} = C_c(Y)$. 
To use discrete optimal transport, we also define two empirical measures $\hat{\mu}_r$ and $\hat{\mu}_c$ based on $X$ and $Y$ as follows:
$\hat{\mu}_r = \frac{1}{n}\sum_{i=1}^{n}\delta_{\bm{x}_i}$ and $\hat{\mu}_c = \frac{1}{m}\sum_{i=1}^{m}\delta_{\bm{y}_i}$. We are now ready to present our method. 

\subsection{Proposed approach}
The main underlying idea of our approach is to use the optimal transportation presented above to find a probabilistic coupling of the empirical measures defined based on rows and columns of a given data matrix. 
More formally, for some fixed $\lambda >0$ we solve the co-clustering problem through the following optimization procedure:
\begin{align}
    \gamma^*_{\lambda} = \argmin_{\gamma \in \Pi(\hat{\mu}_r, \hat{\mu}_c)}\langle M, \gamma\rangle_F - \frac{1}{\lambda}E(\gamma),
    \label{eq1:algo1}
\end{align}

where the matrix $M$ is computed using the Euclidean distance, \ie, $
M_{ij} = \Vert \bm{x}_i - \bm{y}_j \Vert^2$. The elements of the resulting matrix $\gamma^*_{\lambda}$ provides us with the weights of associations between instances and features: similar instances and features correspond to higher values in $\gamma^*_{\lambda}$. Our intuition is to use these weights to identify the most similar sets of rows and columns that should be grouped together to form co-clusters. 


Following \cite{2015-Benamou-Bregman}, this optimization problem can be equivalently rewritten in the following way:
\begin{align*}
    \min_{\gamma \in \Pi(\hat{\mu}_r, \hat{\mu}_c)}\langle M, \gamma\rangle_F - \frac{1}{\lambda}E(\gamma) = \frac{1}{\lambda} \min_{\gamma \in \Pi(\hat{\mu}_r, \hat{\mu}_c)} \text{KL}(\gamma\Vert \xi_{\lambda}),
\end{align*}
where $\xi_{\lambda} = e^{-\lambda M}$ is the Gibbs kernel. 

Finally, we can rewrite the last expression as follows:
\begin{align*}
    \min_{\gamma \in \Pi(\hat{\mu}_r, \hat{\mu}_c)} \text{KL}(\gamma\Vert \xi_{\lambda}) = \min_{\gamma \in \mathcal{C}} \text{KL}(\gamma\Vert \xi_{\lambda}),
\end{align*}
where $\mathcal{C} = \mathcal{C}_1 \cap \mathcal{C}_2$ is the intersection of closed convex subsets given by $\mathcal{C}_1 = \{\gamma \in \mathbb{R}^{d \times d} \vert \gamma \bm{1} = \hat{\mu}_r\}$ and $\mathcal{C}_2 = \{\gamma \in \mathbb{R}^{d \times d} \vert \gamma^T \bm{1} = \hat{\mu}_c\}$.
The solution of the entropy regularized optimal transport can be obtained using Sinkhorn-Knopp algorithm and has the following form \cite{2015-Benamou-Bregman}:
\begin{align}
    \gamma^*_{\lambda} = \text{diag}(\bm{\alpha})\xi_{\lambda} \text{diag}(\bm{\beta}),
    \label{eq:sk}
\end{align}
where $\bm{\alpha}$ and $\bm{\beta}$ are the scaling coefficients of the Gibbs kernel $\xi_{\lambda}$.

In what follows, we show that under some plausible assumptions, we can interpret these two vectors as approximated rows and columns probability density functions. 

\subsection{Connection to variational inference}
In order to justify our approach from the theoretical point of view, we first explain how the obtained solution $\gamma^*$ can be used for co-clustering. As mentioned in \cite{Dhillon:2003:IC:956750.956764} and later in \cite{Banerjee:2007:GME:1314498.1314563}, the co-clustering can be seen as a density estimation problem where the goal is to approximate the real density $p(X,Y)$ by a simpler one depending on the obtained co-clustering in a way that it preserves the loss in the mutual information given by $I(X,Y) - I(\hat{X}, \hat{Y})$ where $I(X,Y) = \int_{XY} p(x,y) \log \frac{p(x, y)}{p(x)p(y)} dxdy$ is the mutual information. 
This quantity is further shown to be equal to the Kullback-Leibler divergence between the original distribution $p(X,Y)$ and $q(X,Y)$ where the latter has the following form:
\begin{align*}
    q(x,y) = p(y\vert \hat{y}) p(\hat{x}, \hat{y}) p(x\vert \hat{x}).
\end{align*}
From this point, one may instantly see that the solution of the optimal transport problem $\gamma^*$ has a very similar form as it also represents the joint probability distribution that approximates the original probability distribution $p(x,y)$ given by the Gibbs measure $\xi_{\lambda}$ and also factorizes into three terms. The most important difference, however, lies in the asymmetry of the KL divergence: while \cite{Dhillon:2003:IC:956750.956764} and \cite{Banerjee:2007:GME:1314498.1314563} concentrate on minimizing $\text{KL}(p(X,Y) \Vert q(X,Y))$, our idea is different and consists in minimizing $\text{KL}(q(X,Y) \Vert p(X,Y))$. This approach is known in the literature as the variational inference \cite{Bishop:2006:PRM:1162264} and exhibits a totally different behaviour compared to the minimization of $\text{KL}(p(X,Y) \Vert q(X,Y))$. As shown by \cite{Bishop:2006:PRM:1162264}, in variational inference the estimated distribution $q(X,Y)$ concentrates on the modes of data and remains compact, while the minimizer of $ \text{KL}(p(X,Y) \Vert q(X,Y))$ tends to cover the whole surface of the original density and to overestimate its support. As $X$, $Y$ and $\hat{X}$ and $\hat{Y}$ represent the observed and unobserved variables, respectively, the natural goal is to try to estimate the distribution $p(X, Y \vert \hat{X},\hat{Y})$ of the data given the obtained co-clusters by the simpler variational distribution $q(X,Y)$. 
However, as the maximisation of $p(X, Y \vert \hat{X},\hat{Y})$ is computationally impossible, it is common to introduce a free distribution $q(\cdot,\cdot)$ on the parameters $\widehat{X}$ and $\widehat{Y}$ in order to obtain the following decomposition:
\begin{align*}
    \!\log p(X, Y)\! =\! \mathcal{L}(q(\hat{X},\hat{Y})) \!+\! \text{KL}(q(\hat{X},\hat{Y}) \Vert p(\hat{X},\hat{Y}\vert X, Y)),
\end{align*}
where the lower bound 
$$\mathcal{L}(q(\hat{X},\hat{Y}))  = \int_{\hat{x}\in\hat{\mathcal{X}}}\int_{\hat{y}\in\hat{\mathcal{Y}}} q(\hat{x},\hat{y}) \log \frac{p(x, y, \hat{x},\hat{y})}{q(\hat{x},\hat{y})}d\hat{x}d\hat{y}$$
is maximized when the KL divergence is minimized. 

Now, if we assume that $p(\hat{X},\hat{Y}\vert X, Y)$ follows the Gibbs distribution, i.e. $p(\hat{X},\hat{Y}\vert X, Y ) \propto e^{-\lambda M(x,y)}$, we can consider the original formulation of the regularized optimal transport as the variational inference problem:
\begin{align*}
    \min_{q}\text{KL}(q(\hat{X},\hat{Y}) \Vert p(\hat{X},\hat{Y}\vert X, Y)) = \min_{\gamma}\text{KL}(\gamma\Vert \xi_{\lambda}),
\end{align*}
where the optimal coupling $\gamma$ equals to the estimated joint probability $q(\hat{X},\hat{Y})$.

At this point, we know that the coupling matrix can be seen as an approximation to the original unknown posterior density function but the question how one can use it to obtain the clustering of rows and columns has not been answered yet. In order to solve the variational inference problem, it is usually assumed that the variables $\hat{x}, \hat{y}$ are independent and thus the variational distribution $q(\hat{x},\hat{y})$ factorizes as $q(\hat{x},\hat{y}) = q(\hat{x})q(\hat{y})$. This assumption, however, goes against the whole idea of co-clustering that relies on the existence of a deep connection between these two variables. 

To this end, we propose to consider the factorization of $q(\hat{x},\hat{y})$ that has the following form
$$q(\hat{x},\hat{y}) = q(x) q(\hat{x}, \hat{y} \vert x,y) q(y).$$
This particular form follows the idea of structured stochastic variational inference proposed in \cite{DBLP:conf/aistats/HoffmanB15} where a term depicting the conditional distribution between hidden and observed variables is added to the fully factorized traditional setting presented above. As stated in \cite{DBLP:conf/aistats/HoffmanB15}, this term allows arbitrary dependencies between observed and hidden variables which can increase the fidelity of the approximation. 

Following \cite{Bishop:2006:PRM:1162264}, the optimal estimated densities $q(x)$ and $q(y)$ are controlled by the direction of the smallest variance of $p(x)$ and $p(y)$ respectively. Furthermore, $q(x)$ and $q(y)$ are proportional to the joint densities $p(\hat{y}, y)$ and $p(\hat{x}, x)$, \ie, $q(x) \propto p(\hat{y}, y)$ and $q(y) \propto p(\hat{x}, x)$. Bearing in mind the equivalence between $\gamma^*_{\lambda}$ and $q(\hat{x},\hat{y})$, this brings us to the following important conclusions: (1) the matrices $\text{diag}(\bm{\alpha})$ and $\text{diag}(\bm{\beta})$ can be seen as the approximated densities $p(\hat{Y}, Y)$ and $p(\hat{X}, X)$; (2) vectors $\bm{\alpha}$ and $\bm{\beta}$ represent the approximated densities $p(X)$ and $p(Y)$ obtained by summing $X$ and $Y$ out of $p(\hat{X}, X)$ and $p(\hat{Y}, Y)$, respectively. 

According to \cite{Laird:78}, the non-parametric estimate of the mixing distribution is a piecewise step function where the number of steps depend on the number of components in the mixture. In the cluster analysis, we can assume that $p(X)$ and $p(Y)$ consist of $g$ and $m$ components, respectively. Then, our goal is to detect these steps based on the estimates given by $\bm{\alpha}$ and $\bm{\beta}$ to obtain the desired partitions. 

\subsection{Kernelized version and Gromov-Wasserstein distance}
In this part, we introduce the kernelized version of our method and compare it to the original formulation of our algorithm. In order to proceed, we first define two similarity matrices $K_r \in \mathbb{R}^{n\times n} $ and $K_c \in \mathbb{R}^{d \times d}$ associated to empirical measures $\hat{\mu}_r, \hat{\mu}_c$ thus forming metric-measure spaces as in \cite{journals/focm/Memoli11}. Matrices $K_r$ and $K_c$ are defined by calculating the pairwise distances or similarities between rows and columns, respectively, without the restriction of them being positive or calculated based on a proper distance function satisfying the triangle inequality. The entropic Gromov-Wasserstein discrepancy in this case is defined as follows \cite{DBLP:conf/icml/PeyreCS16}:
\begin{align*}
    &\text{GW}(K_r,K_c,\hat{\mu}_r, \hat{\mu}_c) = \min_{\gamma \in \Pi_{\hat{\mu}_r,\hat{\mu}_c}}\Gamma_{K_r,K_c}(\gamma) -\lambda E(\gamma) \\
    & = \min_{T \in \Pi_{\hat{\mu}_r,\hat{\mu}_c}} \sum_{i,j,k,l} L(K_{r_{i,j}},K_{c_{k,l}})\gamma_{i,j}\gamma_{k,l} -\lambda E(\gamma).
\end{align*}
where $\gamma$ is a coupling matrix between two similarity matrices and $L:\mathbb{R} \times \mathbb{R} \rightarrow \mathbb{R}_{+}$ is an arbitrary lost-function, usually the quadratic-loss or Kullback-Leibler divergence.

Based on this definition, one may define the problem of the entropic Gromov-Wasserstein barycenters for similarity or distance matrices $K_r$ and $K_c$ as follows:
\begin{align}
    \min_{K,\gamma_r,\gamma_c} \sum_{i=\{r,c\}} \varepsilon_i \Gamma_{K,K_i}(\gamma_i) -\lambda E(\gamma_i)
    \label{eq:algo2}
\end{align}
where $K$ is the computed barycenter and $\gamma_r \in \Pi_{\hat{\mu},\hat{\mu}_r}$, $\gamma_c \in \Pi_{\hat{\mu},\hat{\mu}_c}$ are the coupling matrices that align it with $K_r$ and $K_c$, respectively. $\varepsilon_i$ are the weighting coefficients summing to one, \ie, $\sum_{i=\{r,c\}} \varepsilon_i = 1$ that determine our interest in more accurate alignment between $K_r$ and $K$ or $K_c$ and $K$.

The intuition behind this optimization procedure for co-clustering with respect to original formulation given in (\ref{eq1:algo1}) is the following: while in (\ref{eq1:algo1}) we align rows with columns directly, in (\ref{eq:algo2}) our goal is to do it via an intermediate representation given by the barycenter $K$ that is optimally aligned with both $K_r$ and $K_c$. In this case, we obtain the solutions $\gamma_r$ and $\gamma_c$ that, similar to (\ref{eq:sk}), can be decomposed as follows:
\begin{align*}
    \gamma_r^* = \text{diag}(\bm{\alpha_r})\xi_r\text{diag}(\bm{\beta_r}),\,
    \gamma_c^* = \text{diag}(\bm{\alpha_c})\xi_c\text{diag}(\bm{\beta_c})
    \label{eq:skgw}
\end{align*}
where $\xi_r = e^{-\lambda M_r}$ and $\xi_c = e^{-\lambda M_c}$ are Gibbs kernels calculated between the barycenter and row and column similarity matrices using any arbitrary loss-function $L$ as explained before. Finally, based on the analysis presented above, we further use vectors $\bm{\beta_r}$ and $\bm{\beta_c}$ to derive row and column partitions. 

\subsection{Detecting the number of clusters}\label{sec:jump_detection}

In order to detect the steps (or jumps) in the approximated marginals, we propose to adapt a procedure introduced in \cite{MatMei} for multiscale denoising of piecewise smooth signals. This method is of particular interest for us as it determines the significant jumps in the vectors $\bm{\alpha}$ and $\bm{\beta}$ without knowing their number and location, nor a specific threshold to decide the significance of a jump. As the proposed procedure deals with non-decreasing functions, we first sort the values of $\bm{\alpha}$ and $\bm{\beta}$ in the ascending order.
Since the procedure is identical for both vectors, we only describe it for the vector $\bm{\alpha}$. 

We consider that the elements $\{\alpha_i\}_{i=1}^n$ of $\bm{\alpha}$ are the local averages of a piecewise continuous function $v:[0,1[ \subset\R \rightarrow \R$ on the intervals $I^{n}_{i} =[i/n,(i+1)/n[$ defined by the uniform subdivision of step $1/n$ of the interval $[0,1[$. More precisely:
$
\alpha^n_i = n\int_{I^{n}_{i}} v(t)dt, i=0,\ldots,n-1 
$. 
The detection strategy is based on the following cost function: 
$
F(I^{n}_{i})= \sum_{l=i-1}^i | \alpha^n_{l+1}- \alpha^n_l|
$
defined for each interval.
Therefore, we get the list of the interval suspicious to contain a jump for the subdivision of order $n$ as follows:
\begin{equation*}
\label{listn}
L^n=\{i^*; i^*=\argmax_i F(I^{n}_{i})\}.
\end{equation*}
This detection should be refined in order to get only significant jumps in our vector $\bm{\alpha}.$ To this end we use the  multiscale representation of $\bm{\alpha}$ as in \cite{HartenSR} and we perform this detection on each scale. On the first scale, we get a coarse version of $\bm{\alpha}$ by averaging: 
$$
\alpha^{n/2}_i = \frac 1 2 (\alpha^{n}_{2i} + \alpha^{n}_{ 2i+1}),i=0,\ldots,n/2-1.
$$
Now, by considering the coarse version of $\bm{\alpha}$,
we obtain a second list $L^{n/2}$ of suspicious intervals as before.
After that, these two lists merge in the list $L_{\text{jumps}}$ as follows: a  jump will be considered in the interval 
$I^{n}_{2i}$ or   $I^{n}_{2i+1}$ if the interval $I^{n/2}_{i}$ is also detected as suspicious at the coarse scale. 
This procedure is iterated $[\log_2 n]$ times and a jump is observed if a chain of detection exists from fine to coarse scales. Finally, the number of clusters is obtained by $g = \vert L_{\text{jumps}} \vert +1$.

\subsection{Algorithmic implementation}

We now briefly summarize the main steps of both {\CCOT} and {\CCOTGW} methods and discuss their peculiarities with respect to each other. The pseudocode of both approaches in Matlab are presented in \Cref{algo} and \Cref{algo2}, respectively.

\paragraph{{\CCOT}} First step of our algorithm consists in calculating the cost matrix $M$ and using it to obtain the optimal coupling matrix $\gamma^*_{\lambda}$ by applying the regularized optimal transport. In order to calculate $M$, row and column instances should both lie in a space of the same dimension. This condition, however, is verified only if the matrix $\mathcal{A}$ is squared which occurs rarely in the real-world applications. To overcome this issue, we first subsample the original data set $\mathcal{A}$ in a way that allows us to equalize the number of rows and columns and operate with two sets of the same dimension. If we assume that $n>d$ then this new reduced data set is denoted by $D \in \mathbb{R}^{d \times d}$. We repeat the sampling procedure until every individual is picked at least once. 

The next step is to perform for each $i=1,\dots,n_s$ the jump detection on the sorted vectors $\bm{\alpha}_i$ and $\bm{\beta}_i$ to obtain two lists of the jumps locations $L_{\text{jumps}}^{\alpha_i}$ and $L_{\text{jumps}}^{\beta_i}$ and to define the number of row and column clusters $g$ and $m$. By using them, we obtain the resulting row partition:
$$    C_r^{i}(\bm{x}_r)=\left\{
                \begin{array}{ll}
                  1, \;\ r \leq L_{\text{jumps}}^{\alpha_i}(1)\\
                  \dots\\
                  k,\; \ L_{\text{jumps}}^{\alpha_i}(k-1) <r\leq L_{\text{jumps}}^{\alpha_i}(k)\\
                  \dots\\
                  \vert L_{\text{jumps}}^{\alpha_i}\vert + 1, \; \ r >  L_{\text{jumps}}^{\alpha_i}(\vert L_{\text{jumps}}^{\alpha_i}\vert).
                \end{array}
              \right.
$$
The partition for columns $C_c^{i}(\bm{y}_c)$ is obtained in the same way. 
Finally, we apply the majority vote over all samples partitions to obtain $C_r$ and $C_c$.
Regarding complexity, both Sinkhorn-Knopp algorithm used to solve the regularized optimal transport \cite{Knight:2008:SAC:1404637.1404647} and the proposed jump detection techniques are known to converge at the linear rate multiplied by the number of samples, \ie, $\mathcal{O}(n_sd)$. On the other hand, the calculation of modes of the clustering obtained on the generated samples for both features and data instances has the complexity $\mathcal{O}(n_s(n+d))$. In the end, the complexity of the whole algorithm is $\mathcal{O}(n_s(n+d))$. We also note that in the real-world applications, we usually deal with scenarios where $n\gg d$ (``big data") or $d\ll n$ (``small" data) thus reducing the overall complexity to $\mathcal{O}(n_sn)$ and $\mathcal{O}(n_sd)$, respectively. This makes our approach even more computationally attractive.

\begin{algorithm}[!ht]
\begin{small}
 \SetKwInOut{Input}{Input}\SetKwInOut{Output}{Output}
 \Input{$\mathcal{A}$ - data matrix, $\lambda$ - regularization parameter, $n_{s}$  - number of sampling}
 \Output{$C_r, C_c$ - partition matrices for rows and columns, $g, m$ - number of row and column clusters}

 $\mathtt{[n, d] = size(Z)}$\;

 \For{$i = 1$ \KwTo $n_{s}$}{
 $\mathtt{D_i = datasample(Z,d)}$ \;
 
 $\mathtt{M_i \leftarrow pdist2(D_i,D_i^T)}$\;
 
 $\mathtt{[\bm{\alpha}_i, \bm{\beta}_i, \gamma^*] \leftarrow optimal\_transport(M_i,\lambda)}$\;
 
 $\mathtt{[L_{\text{jumps}}^{\alpha_i},C_r^{i},g] \leftarrow jump\_detection(sort(\bm{\alpha}_i))}$ \;
 
 $\mathtt{[L_{\text{jumps}}^{\beta_i},C_c^{i},m] \leftarrow jump\_detection(sort(\bm{\beta}_i))}$ \;
 }

 $\mathtt{C_r \leftarrow mode(C_r^{i})}$\;
 
 $\mathtt{C_c \leftarrow mode(C_c^{i})}$\;
 \caption{Co-clustering through Optimal Transport ({\CCOT})}
 \label{algo}
 \end{small}
\end{algorithm}

\paragraph{{\CCOTGW}} As it can be seen from \Cref{algo2}, {\CCOTGW} allows to overcome the important disadvantage of {\CCOT} that consists in the need to perform sampling to cluster all data objects. On the other hand, the computational complexity of {\CCOT} is only $\mathcal{O}(n_sd)$, while for {\CCOTGW} it scales as $\mathcal{O}(n^2d+d^2n)$. We also note that {\CCOTGW} offers a great flexibility in terms of the possible data representation used at its input. One may easily consider using any arbitrary kernel function to calculate similarity matrices or even learn them beforehand using multiple-kernel learning approaches. 

\begin{algorithm}[!htpb]
\begin{small}
 \SetKwInOut{Input}{Input}\SetKwInOut{Output}{Output}
 \Input{$\mathcal{A}$ - data matrix, $\lambda$ - regularization parameter, $\varepsilon_r$, $\varepsilon_c$ - weights for barycenter calculation}
 \Output{$C_r, C_c$ - partition matrices for rows and columns, $g, m$ - number of row and column clusters}

 $\mathtt{K_r \leftarrow pdist2(Z,Z)}$\;
 
 $\mathtt{K_c \leftarrow pdist2(Z^T,Z^T)}$\;
 
 $\mathtt{[\bm{\beta}_r, \bm{\beta}_c, \gamma^*_r, \gamma^*_c ]\leftarrow gw\_barycenter(K_r,K_c,\lambda,\varepsilon_r, \varepsilon_c)}$\;
 
 $\mathtt{[L_{\text{jumps}}^{\beta_r},C_r,g] \leftarrow jump\_detection(sort(\bm{\beta}_r))}$ \;
 
 $\mathtt{[L_{\text{jumps}}^{\beta_c},C_c,m] \leftarrow jump\_detection(sort(\bm{\beta}_c))}$ \;
 
 \caption{Co-clustering through Optimal Transport with Gromov-Wasserstein barycenters ({\CCOTGW})}
 \label{algo2}
 \end{small}
\end{algorithm}

\begin{table*}[!ht]
\renewcommand\thetable{2}
\caption{Mean ($\pm$ standard-deviation) of the co-clustering error (CCE) obtained for all configurations. ``-" indicates that the algorithm cannot find a partition with the requested number of co-clusters. P-values obtained using the non-parametric test of Wilcoxon \cite{wilcoxon_45} that imply significant differences are printed in bold (significance level of 0.05).
}
\label{aaai17:tab:result}
\begin{center}
\resizebox{0.9\textwidth}{!}{
\begin{tabular}{lccccccccc}
\hline
\multirow{2}{*}{Data set} &\multicolumn{9}{c}{Algorithms}   \\
\cline{2-10}
  & K-means&NMF&DKM&Tri-NMF&GLBM&ITCC&RBC&\CCOT&\CCOTGW\\
\hline
D1&$\mathbf{.018\pm{.003}}$&$.042\pm{.037}$&$.025\pm{.048}$&$.082\pm{.063}$&$.021\pm{.011}$&$.021\pm{.001}$&$.017\pm{.045}$&$.018\pm{.013}$&$\mathbf{.004\pm{.002}}$\\
D2&$.072\pm{.044}$&$.083\pm{.063}$&$.038\pm{.000}$&$.052\pm{.065}$&$.032\pm{.041}$&$.047\pm{.042}$&$.039\pm{.052}$&$.023\pm{.036}$&$\mathbf{.011\pm{.056}}$\\
D3&--&--&$.310\pm{.000}$&--&$.262\pm{.022}$&$.241\pm{.031}$&--&$.031\pm{.027}$&$\mathbf{.008\pm{.001}}$\\
D4&$.126\pm{.038}$&--&$.145\pm{.082}$&--&$.115\pm{.047}$&$.121\pm{.075}$&$.102\pm{.071}$&$.093\pm{.032}$&$\mathbf{.079\pm{.031}}$\\
\hline
\end{tabular}
}
\end{center}
\end{table*}

\section{Experimental evaluations}
In this section, we provide empirical evaluation for the proposed algorithms.
\subsection{Synthetic data}
\begin{figure}[!htpb]
  \begin{center}
    \subfloat[D2]{
      \includegraphics[width=2.6cm, height=2.3cm]{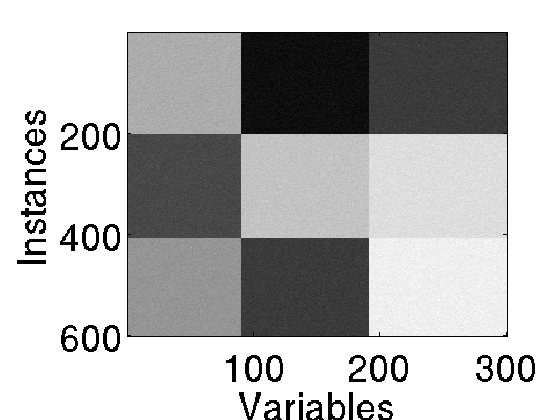}
      \label{sub:original_d1}
      }
    \subfloat[D3]{
      \includegraphics[width=2.6cm, height=2.3cm]{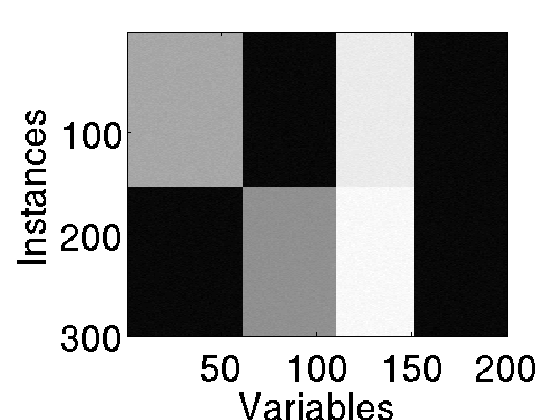}
      \label{sub:reorganised_d3}}
      \subfloat[D4]{
      \includegraphics[width=2.6cm, height=2.3cm]{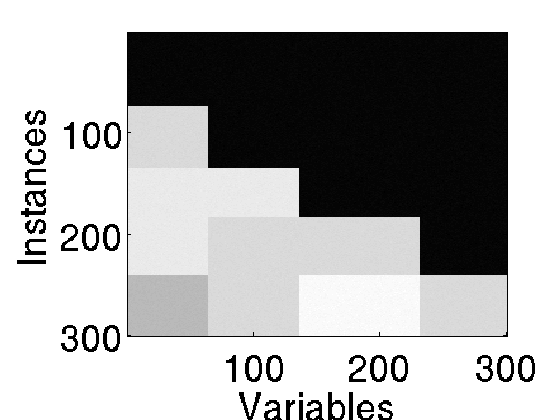}
      \label{sub:reorganised_d2}}
    \caption{D2, D3 and D4 reorganized w.r.t. the true partitions.}
    \label{aaai17:fig:visu_d2}
  \end{center}
\end{figure}

\paragraph{Simulation setting}
We simulate data following the generative process of the Gaussian Latent Block Models (for details see \cite{govaert2013co}) and we consider four scenarios with different number of co-clusters, degree of separation and size. \Cref{aaai17:tab:data_description} and \Cref{aaai17:fig:visu_d2} present the characteristics of theta simulated data sets and their visualization showing the different co-clustering structures.

\begin{table}[H]
\renewcommand\thetable{1}
\caption{Size ($n\times d$), number of co-clusters ($g\times m$), degree of overlapping ([+] for well-separated and [++] for ill-separated co-clusters) and the proportions of co-clusters for simulated data sets.}
\label{aaai17:tab:data_description}
\centering
\resizebox{0.45\textwidth}{!}{
\begin{tabular}{lccccl}
\hline
Data set&$n\times d$&$g\times m$&Overlapping&Proportions\\
\hline
D1&$600\times300$&$3\times 3$&[+]&Equal\\
D2&$600\times300$&$3\times 3$&[+]&Unequal\\
D3&$300\times200$&$2\times 4$&[++]&Equal\\
D4&$300\times300$&$5\times4$&[++]&Unequal\\
\hline
\end{tabular}
}
\end{table}


We use several state-of-the-art co-clustering algorithms as baselines including ITCC \cite{Dhillon:2003:IC:956750.956764}, Double K-Means (DKM)~\cite{rocci_08}, Orthogonal Nonnegative Matrix Tri-Factorizations (ONTMF) \cite{ding_06}, the Gaussian Latent Block Models (GLBM) \cite{NADI08CI,govaert2013co} and Residual Bayesian Co-Clustering (RBC) \cite{shan_10}. 
We also report the results of K-means and NMF, run on both modes of the data matrix, as clustering baseline. To assess the performance of all compared methods, we compute the co-clustering error (CCE) \cite{Patrikainen06} defined as follows:
$$\text{CCE}((\bz,\bw),(\hat{\bz},\hat{\bw}))=e(\bz,\hat{\bz})+ e(\bw,\hat{\bw})-e(\bz,\hat{\bz})\times e(\bw,\hat{\bw}),$$
where $\hat{\bz}$ and $\hat{\bw}$ are the partitions of instances and variables estimated by the algorithm; $\bz$ and $\bw$ are the true partitions and $e(\bz,\hat{\bz})$ (resp. $e(\bw,\hat{\bw})$) denotes the error rate, i.e., the proportion of misclassified instances (resp. features). 

For all configurations, we generate 100 data sets and compute the mean and standard deviation of the CCE over all sets. As all the approaches we compared with are very sensitive to the initialization, we run them $50$ times with random initializations and retain the best result according to the corresponding criterion. RBC is initialized with K-means.
Regarding {\CCOT} we set $n_{s}$ to $1000$ for all configurations except D4 which has the same number of rows and columns, and therefore does not require any sampling. For {\CCOTGW}, we use Gaussian kernels for both rows and columns with $\sigma$ computed as the mean of all pairwise Euclidean distances between vectors \cite{DBLP:conf/nips/KarJ11}. Finally, we let both {\CCOT} and {\CCOTGW} detect automatically the number of co-clusters, while for all other algorithms we set the number of clusters to its true value.

\paragraph{Co-clustering performance}
We report the mean (and standard deviation) of co-clustering errors obtained in \Cref{aaai17:tab:result}. Based on these results, we observe that on D1, which has a clear block structure, all algorithms perform equally well, however {\CCOTGW} gives the best results, closely followed by {\CCOT} and K-means. Regarding D2, D3 and D4, which have more complicated structure than D1, both {\CCOT} and {\CCOTGW} significantly outperform all other algorithms and this difference is all the more important on D3 and D4 where some of the compared algorithms are unable to find a partition with the desired number of clusters.

\begin{figure}[!h]
  \begin{center}
    \subfloat[]{
      \includegraphics[width=0.235\textwidth]{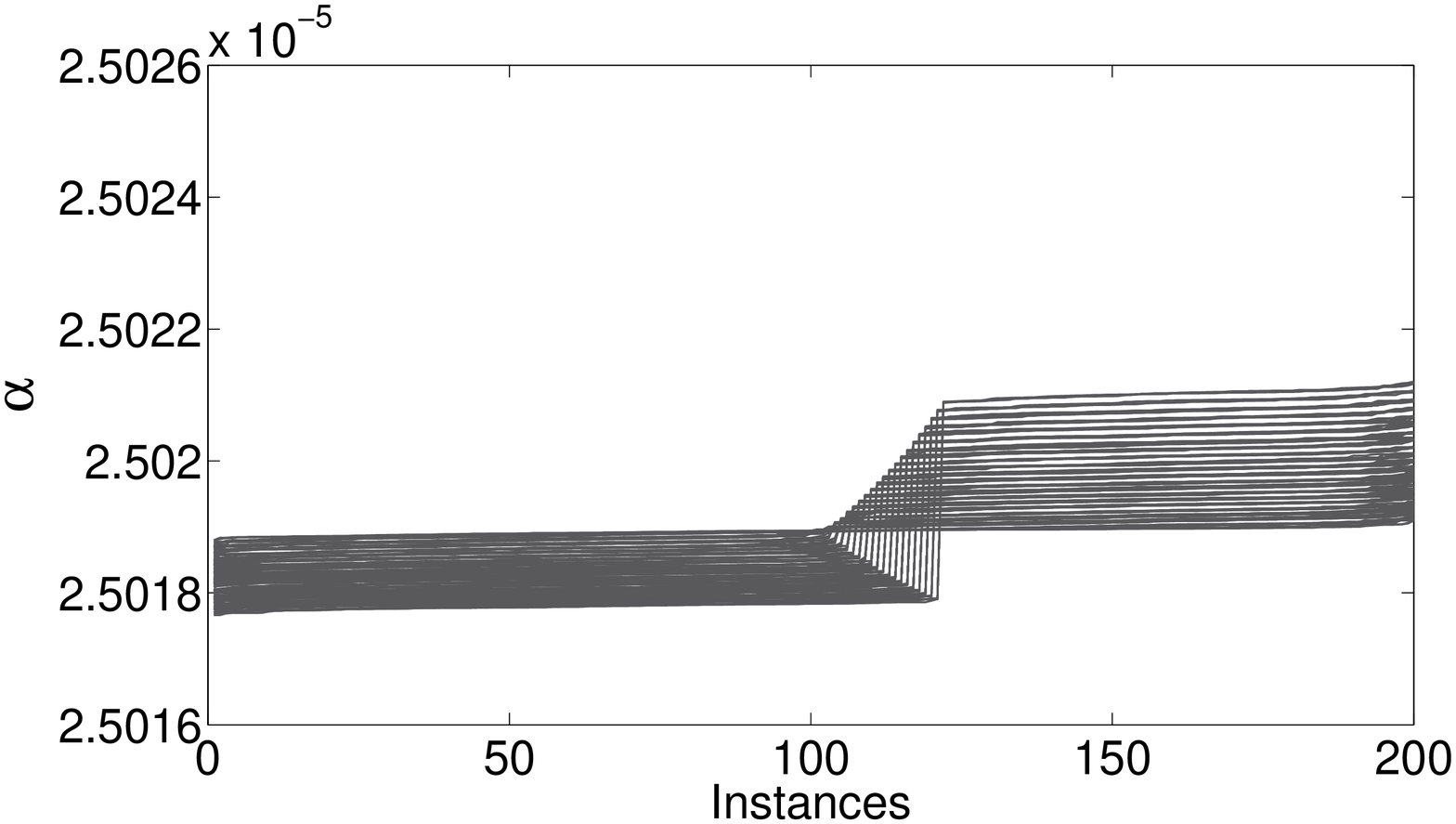}
      \label{sub:alpha_2}
                         }
    \subfloat[]{
      \includegraphics[width=0.235\textwidth]{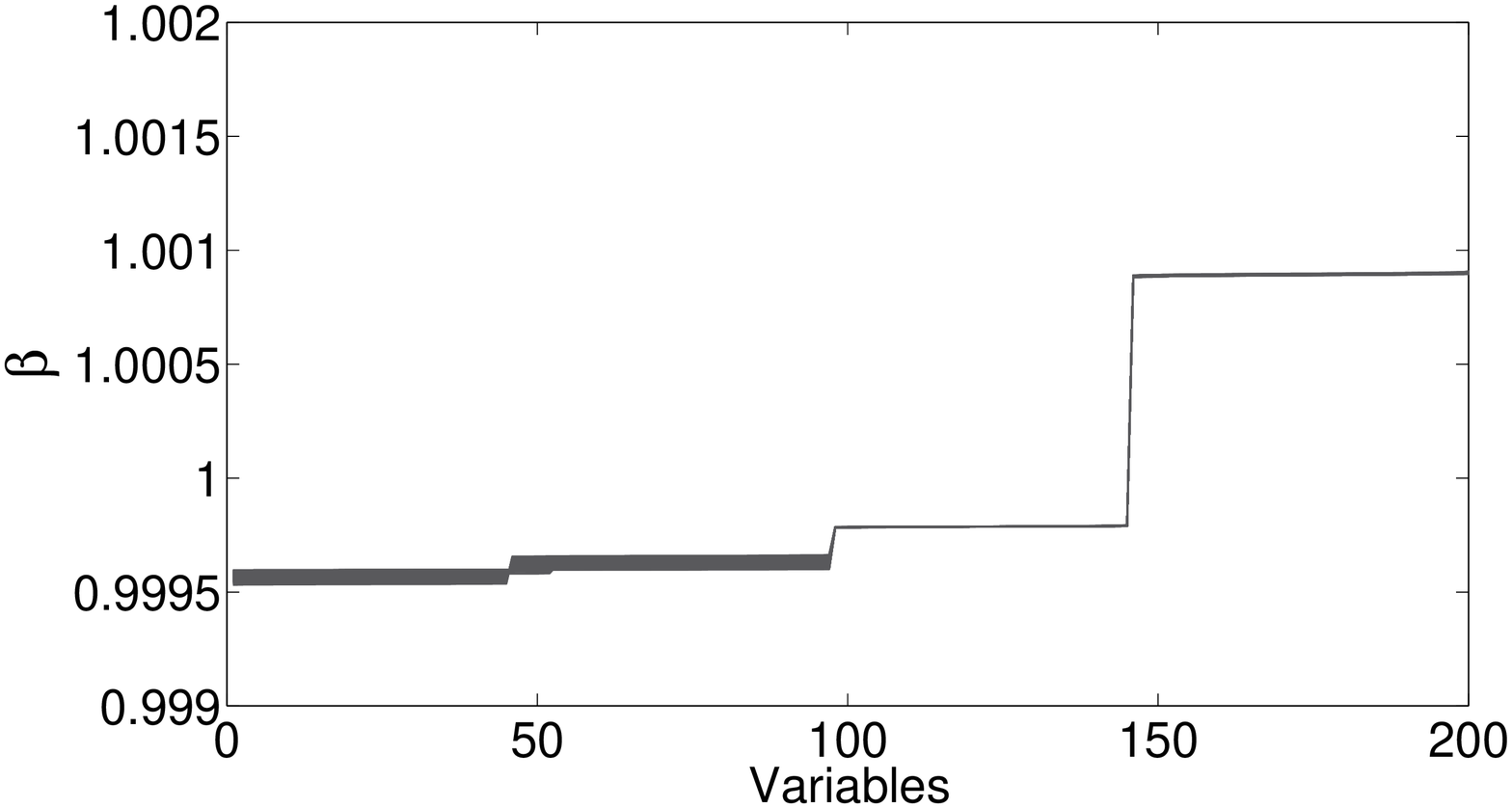}
      \label{sub:beta_4}
                         }\\
       \subfloat[]{
       \scriptsize{
       \begin{tabular}{lcccc}
    \hline
    Data Set& \multicolumn{2}{c}{\CCOT}&\multicolumn{2}{c}{\CCOTGW}\\
    \cline{2-5}
    &$g$&$m$&$g$&$m$\\
    \hline
        D1  &$\mathbf{100}$&$\mathbf{100}$&$98$&$100$\\
        D2  &$\mathbf{83}$&$\mathbf{97}$&$81$&$72$\\
        D3  &$\mathbf{99}$&$\mathbf{98}$&$93$&$97$\\
        D4  &$\mathbf{73}$&$86$&$70$&$\mathbf{87}$\\
    \hline
    \end{tabular}}
      \label{tab:nbcocluster}
                         }  
    \caption{Vectors (a) $\bm{\alpha}$ and (b) $\bm{\beta}$ obtained with {\CCOT} on D3. (c) Number of times {\CCOT} and {\CCOTGW} correctly detect the number of co-clusters ($g$ and $m$) over 100 trials.}
    \label{aaai17:fig:visu_a_b}
  \end{center}
\end{figure}

Furthermore, we argued that one of the strengths of our method is its ability to detect automatically the number of co-clusters by applying a jump detection algorithm on $\bm{\alpha}$ and $\bm{\beta}$. From \Cref{aaai17:fig:visu_a_b} one can observe that the plots of these vectors, obtained with {\CCOT}, with their elements sorted in the ascending order reveal clear steps that correspond to the correct number of clusters and also illustrate their proportions and the degree of overlapping. The same observation is valid for {\CCOTGW}. Both approaches correctly identified the number of clusters in most cases and {\CCOT} is slightly more accurate than {\CCOTGW} when the proportions of co-clusters are unbalanced.

To summarize, {\CCOT} and {\CCOTGW} outperform all the other baselines for the considered data structures and present two important advantages: (1) they do not suffer from the initialization issues, (2) they are able to detect automatically the number co-clusters. 

\subsection{MovieLens}
\paragraph{Data and setting} \MovieL-100K\footnote{https://grouplens.org/datasets/movielens/100k/} is a popular benchmark data set that consists of user-movie ratings, on a scale of one to five, collected from a movie recommendation service gathering 100,000 ratings from 943 users on 1682 movies. In the context of co-clustering, our goal is to find homogeneous subgroups of users and films in order to further recommend previously unseen movies that were highly rated by the users from the same group.

We set the regularization parameters for {\CCOT} and {\CCOTGW} using the cross-validation; the number of samplings for {\CCOT} is set to $500$ (as the dimensions of the data set are quite balanced); the weights for the barycenter in {\CCOTGW} are set to $\varepsilon=(0.5,0.5)$.

\begin{figure}[!ht]
  \begin{center}
    \subfloat[]{
      \includegraphics[width=0.21\textwidth]{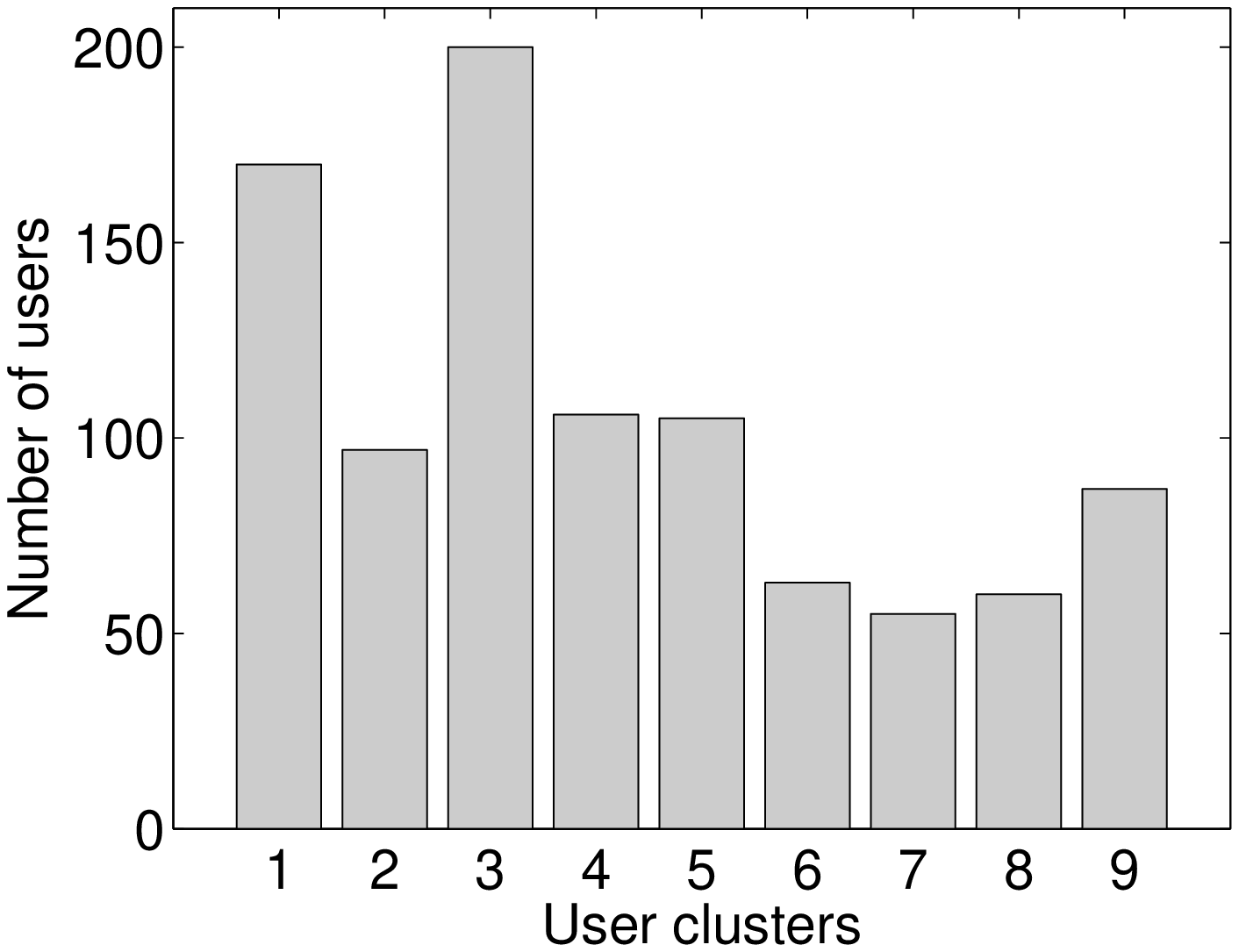}
      \label{sub:user}
                         }
    \subfloat[]{
      \includegraphics[width=0.21\textwidth]{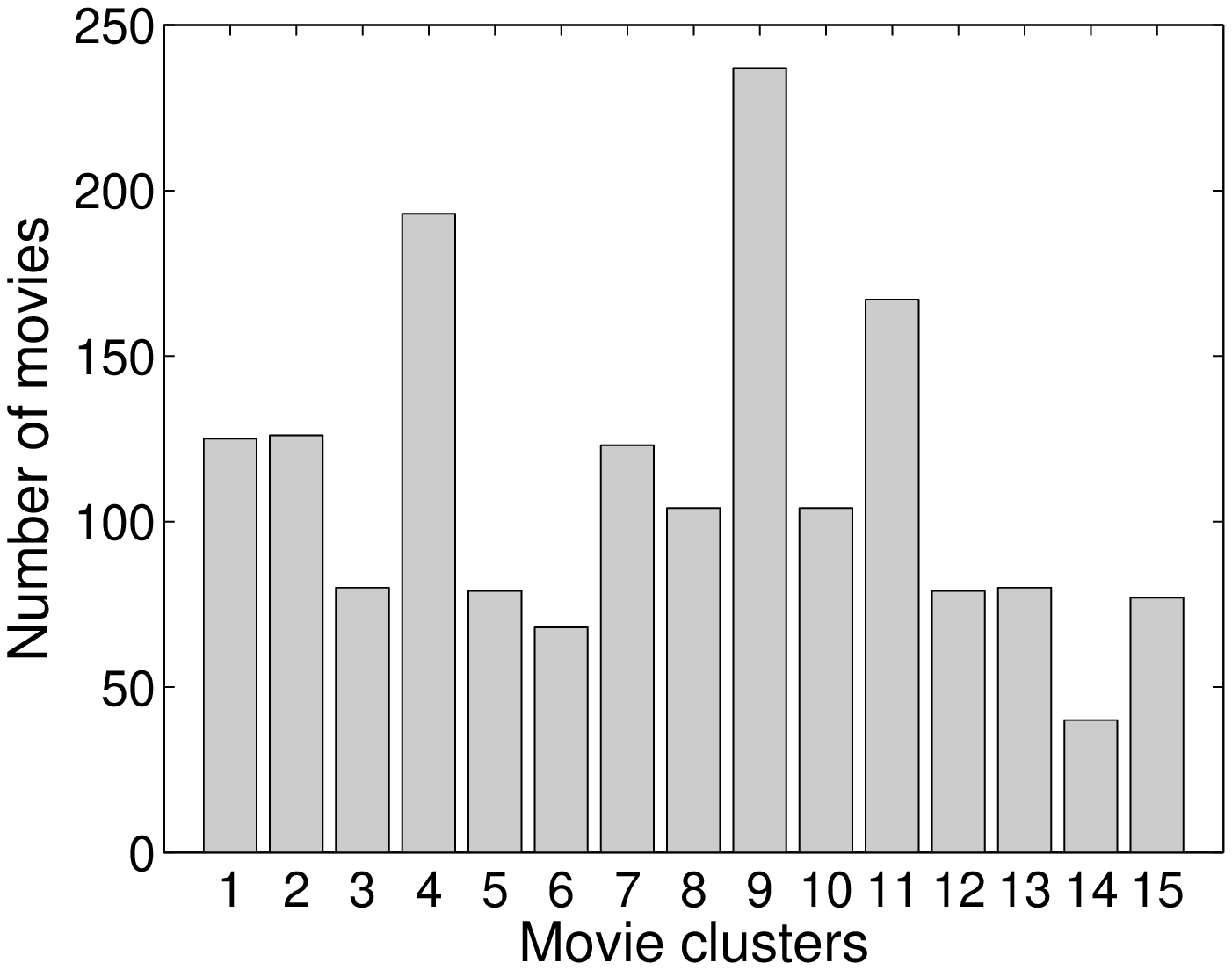}
      \label{sub:movie}
                         }
    \caption{Distribution of the number of (a) users and (b) movies across the clusters obtained with {\CCOTGW}.}
    \label{fig:distribution-ml}
  \end{center}
\end{figure}

\paragraph{Results}
In what follows we only present figures and results obtained by {\CCOTGW} as both algorithms return the same number of blocks and the partitions are almost identical (with a normalized mutual information between partitions above $0.8$). 
{\CCOTGW} automatically detects a structure consisting of $9\times15$ blocks, that corresponds to 9 user clusters and 15 movie clusters. From Figure \ref{fig:distribution-ml}, one can observe that the users and the movies are almost equally distributed across clusters, except for two user and three movie clusters which have a larger size than others. 

\begin{figure}[!ht]
\begin{center}
\subfloat[]{
      \includegraphics[width=0.19\textwidth]{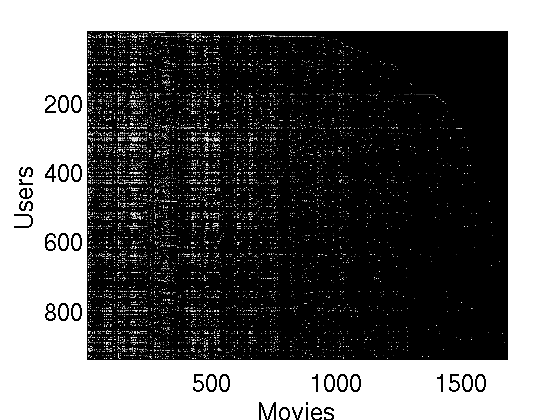}
      \label{sub:original-ml}
                         }
    \subfloat[]{
      \includegraphics[width=0.19\textwidth]{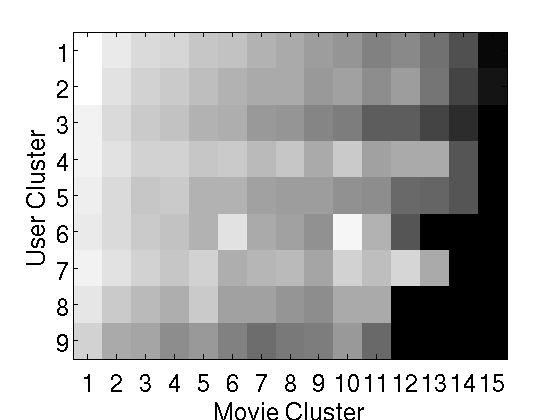}
      \label{sub:summary-ml}
                         }
    \caption{(a) {\MovieL} matrix; (b) the matrix summarized by the mean rating (0 ratings are excluded) for each block obtained with {\CCOTGW}. Darker shades indicate lower values.}
    \label{fig:summ-ml}
\end{center}
\end{figure}

Figure \ref{fig:summ-ml} shows the original data set as well as a summarized version where each block is represented by its mean rating value (the lighter the block, the higher the ratings), revealing a structure into homogeneous groups. One can observe that the first movie cluster consists of films for which all users agree on giving high ratings (most popular movies) while the last movie cluster consists of the movies with very low ratings. We also report the 5 best rated movies in those two clusters in Table \ref{tab:top5}. One can easily see that popular movies, such that both Star Wars episodes are in M1 while M5 is composed of movies that were less critically acclaimed.

\begin{table}[htpb]
\caption{Top 5 of movies in clusters M1 and M15.}
\label{tab:top5}
\centering
\resizebox{0.45\textwidth}{!}{
\begin{tabular}{cc}
\hline
M1&M15\\
\hline
Star Wars (1977)&Amytiville: A New Generation (1993)\\
The Lion King (1994)&Amytiville: It's About Time (1992)\\
Return of the Jedi (1983)&Ninjas: High Noon at Mega Mountain (1998)\\
Contact (1997)&Sudden Manhattan (1996)\\
Raiders of the lost ark (1981)&Dream Man (1995)\\
\hline
\end{tabular}
}
\end{table}

We can make similar observations for the interpretation of user clusters. For instance, the last two user clusters include users that tend to give less good ratings to movies than the average population. Also, we note that block $(6,10)$ corresponds to users who liked movies from M10 better than the rest of the users. These observations are also very similar to the results reported by \cite{Banerjee:2007:GME:1314498.1314563}, where the authors proposed a detailed study of a $10\times20$ blocks structure for this data set. Additional results can be found in the Supplementary material.


\section{Conclusions and future perspectives}
In this paper we presented a novel approach for co-clustering based on the entropy regularized optimal transport. Our method is principally different from other co-clustering methods and consists in finding a probabilistic coupling of the empirical measures defined based on the data instances and features. We showed how this procedure can be seen as the variational inference problem and that the inferred distribution can be used to obtain the row and feature partitions. The resulting algorithm is not only more accurate than other state-of-the-art methods but also fast and capable of automatically detecting the number of co-clusters. We also presented an extended version of our algorithm that makes use of the optimal transportation distance defined on similarity matrices associated to the rows' and columns' empirical measures.

In the future, our work can be continued in multiple directions. First, we would like to extend our method in order to deal with the online setting where the goal is to classify a new previously unseen observation without the need to do the co-clustering of the data set that includes it. This can be done using a recent approach proposed in \cite{DBLP:conf/nips/PerrotCFH16} that allows to update the learned coupling matrix using the out-of-sample observations without recomputing it using all the data. We believe that this extension will make our algorithm attractive for the exploitation in real-time industrial recommendation systems due to its computational efficiency. We would also like to study the generalization properties of our algorithm in a spirit similar to the results obtained in \cite{DBLP:journals/tit/MaurerP10}. This latter work presents a rare case where the generalization bounds are derived for some famous unsupervised learning algorithms.

\section*{Acknowledgements} 
This work has been supported by the ANR project COCLICO, ANR-12-MONU-0001.
\bibliography{_aaai}

\begin{thebibliography}{46}
\providecommand{\natexlab}[1]{#1}
\providecommand{\url}[1]{\texttt{#1}}
\expandafter\ifx\csname urlstyle\endcsname\relax
  \providecommand{\doi}[1]{doi: #1}\else
  \providecommand{\doi}{doi: \begingroup \urlstyle{rm}\Url}\fi

\bibitem[Banerjee et~al.(2007)Banerjee, Dhillon, Ghosh, Merugu, and
  Modha]{Banerjee:2007:GME:1314498.1314563}
Banerjee, Arindam, Dhillon, Inderjit, Ghosh, Joydeep, Merugu, Srujana, and
  Modha, Dharmendra~S.
\newblock A generalized maximum entropy approach to bregman co-clustering and
  matrix approximation.
\newblock \emph{Journal of Machine Learning Research}, 8:\penalty0 1919--1986,
  December 2007.

\bibitem[Benamou et~al.(2015)Benamou, Carlier, Cuturi, Nenna, and
  Peyr{\'e}]{2015-Benamou-Bregman}
Benamou, Jean-David, Carlier, Guillaume, Cuturi, Marco, Nenna, Luca, and
  Peyr{\'e}, Gabriel.
\newblock {Iterative Bregman Projections for Regularized Transportation
  Problems}.
\newblock \emph{{SIAM Journal on Scientific Computing}}, 2\penalty0
  (37):\penalty0 A1111--A1138, 2015.

\bibitem[Bishop(2006)]{Bishop:2006:PRM:1162264}
Bishop, Christopher~M.
\newblock \emph{Pattern Recognition and Machine Learning (Information Science
  and Statistics)}.
\newblock Springer-Verlag New York, Inc., 2006.

\bibitem[Cheng et~al.(2008)Cheng, Law, Siu, and Liew]{cheng_08}
Cheng, Kin-On, Law, Ngai-Fong, Siu, Wan-Chi, and Liew, Alan~W.
\newblock Identification of coherent patterns in gene expression data using an
  efficient biclustering algorithm and parallel coordinate visualization.
\newblock \emph{BMC Bioinformatics}, 9:\penalty0 210, 2008.

\bibitem[Courty et~al.(2014)Courty, Flamary, and Tuia]{courty:hal-01018698}
Courty, Nicolas, Flamary, R{\'e}mi, and Tuia, Devis.
\newblock {Domain adaptation with regularized optimal transport}.
\newblock In \emph{{Proceedings ECML/PKDD 2014}}, pp.\  1--16, 2014.

\bibitem[Cuturi(2013)]{conf/nips/Cuturi13}
Cuturi, Marco.
\newblock Sinkhorn distances: Lightspeed computation of optimal transport.
\newblock In \emph{Proceedings NIPS}, pp.\  2292--2300, 2013.

\bibitem[Cuturi \& Avis(2014)Cuturi and Avis]{Cuturi:2014:GML:2627435.2627452}
Cuturi, Marco and Avis, David.
\newblock Ground metric learning.
\newblock \emph{Journal of Machine Learning Research}, 15\penalty0
  (1):\penalty0 533--564, 2014.

\bibitem[Cuturi \& Doucet(2014)Cuturi and Doucet]{DBLP:conf/icml/CuturiD14}
Cuturi, Marco and Doucet, Arnaud.
\newblock Fast computation of wasserstein barycenters.
\newblock In \emph{Proceedings ICML}, pp.\  685--693, 2014.

\bibitem[Deodhar \& Ghosh(2010)Deodhar and Ghosh]{deodhar10}
Deodhar, M. and Ghosh, J.
\newblock Scoal: A framework for simultaneous co-clustering and learning from
  complex data.
\newblock \emph{ACM Transactions on Knowledge Discovery from Data}, 4\penalty0
  (3):\penalty0 1--31, 2010.

\bibitem[Dhillon et~al.(2003{\natexlab{a}})Dhillon, Mallela, and
  Kumar]{dhillon_03}
Dhillon, Inderjit~S., Mallela, Subramanyam, and Kumar, Rahul.
\newblock A divisive information theoretic feature clustering algorithm for
  text classification.
\newblock \emph{Journal of Machine Learning Research}, 3:\penalty0 1265--1287,
  2003{\natexlab{a}}.

\bibitem[Dhillon et~al.(2003{\natexlab{b}})Dhillon, Mallela, and
  Modha]{Dhillon:2003:IC:956750.956764}
Dhillon, Inderjit~S., Mallela, Subramanyam, and Modha, Dharmendra~S.
\newblock Information-theoretic co-clustering.
\newblock In \emph{Proceedings ACM SIGKDD}, pp.\  89--98, 2003{\natexlab{b}}.

\bibitem[Ding et~al.(2006)Ding, Li, Peng, and Park]{ding_06}
Ding, C., Li, T., Peng, W., and Park, H.
\newblock Orthogonal nonnegative matrix tri-factorizations for clustering.
\newblock In \emph{Proceedings ACM SIGKDD}, pp.\  126--135, 2006.

\bibitem[George \& Merugu(2005)George and Merugu]{merugu_05}
George, T. and Merugu, S.
\newblock A scalable collaborative filtering framework based on co-clustering.
\newblock In \emph{Proceedings ICDM}, pp.\  625--628, 2005.

\bibitem[Govaert \& Nadif(2013)Govaert and Nadif]{govaert2013co}
Govaert, G. and Nadif, M.
\newblock \emph{Co-clustering}.
\newblock John Wiley \& Sons, 2013.

\bibitem[Hanisch et~al.(2002)Hanisch, Zien, Zimmer, and
  Lengauer]{citeulike:350992}
Hanisch, Daniel, Zien, Alexander, Zimmer, Ralf, and Lengauer, Thomas.
\newblock {Co-clustering of biological networks and gene expression data}.
\newblock \emph{BMC Bioinformatics}, 18\penalty0 (suppl 1):\penalty0 145--154,
  2002.

\bibitem[Harten(1989)]{HartenSR}
Harten, Amiram.
\newblock Eno schemes with subcell resolution.
\newblock \emph{Journal of Computational Physics}, 83:\penalty0 148--184, 1989.

\bibitem[Hartigan(1972)]{hartigan-direct-clustering-data-1972}
Hartigan, J.~A.
\newblock {Direct Clustering of a Data Matrix}.
\newblock \emph{Journal of the American Statistical Association}, 67\penalty0
  (337):\penalty0 123--129, 1972.

\bibitem[Hoffman \& Blei(2015)Hoffman and Blei]{DBLP:conf/aistats/HoffmanB15}
Hoffman, Matthew~D. and Blei, David~M.
\newblock Stochastic structured variational inference.
\newblock In \emph{{Proceedings AISTATS}}, volume~38, pp.\  361--369, 2015.

\bibitem[I.~Abraham \& Carlier(2016)I.~Abraham and Carlier]{abraham_2015}
I.~Abraham, R.~Abraham, M.~Bergounioux and Carlier, G.
\newblock Tomographic reconstruction from a few views: a multi-marginal optimal
  transport approach.
\newblock \emph{Applied Mathematics and Optimization}, pp.\  1--19, 2016.

\bibitem[Irpino et~al.(2014)Irpino, Verde, and De~Carvalho]{Irpino:2014}
Irpino, Antonio, Verde, Rosanna, and De~Carvalho, Francisco de~A.T.
\newblock Dynamic clustering of histogram data based on adaptive squared
  wasserstein distances.
\newblock \emph{Expert Systems with Applications}, 41\penalty0 (7):\penalty0
  3351--3366, 2014.

\bibitem[Kantorovich(1942)]{kantorovich}
Kantorovich, Leonid.
\newblock On the translocation of masses.
\newblock In \emph{C.R. (Doklady) Acad. Sci. URSS(N.S.)}, volume 37(10), pp.\
  199--201, 1942.

\bibitem[Kar \& Jain(2011)Kar and Jain]{DBLP:conf/nips/KarJ11}
Kar, Purushottam and Jain, Prateek.
\newblock Similarity-based learning via data driven embeddings.
\newblock In \emph{{NIPS}}, pp.\  1998--2006, 2011.

\bibitem[Keribin et~al.(2015)Keribin, Brault, Celeux, and
  Govaert]{KeribinBCG15}
Keribin, Christine, Brault, Vincent, Celeux, Gilles, and Govaert, G{\'{e}}rard.
\newblock Estimation and selection for the latent block model on categorical
  data.
\newblock \emph{Statistics and Computing}, 25\penalty0 (6):\penalty0
  1201--1216, 2015.

\bibitem[Knight(2008)]{Knight:2008:SAC:1404637.1404647}
Knight, Philip~A.
\newblock The sinkhorn-knopp algorithm: Convergence and applications.
\newblock \emph{SIAM Journal on Matrix Analysis and Applications}, 30\penalty0
  (1):\penalty0 261--275, March 2008.

\bibitem[Laird(1978)]{Laird:78}
Laird, N.
\newblock Nonparametric maximum likelihood estimation of a mixing distribution.
\newblock \emph{Journal of the American Statistical Association}, 73:\penalty0
  805--811, 1978.

\bibitem[Matei \& Meignen(2012)Matei and Meignen]{MatMei}
Matei, Basarab and Meignen, Sylvain.
\newblock Nonlinear cell-average multiscale signal representations: Application
  to signal denoising.
\newblock \emph{Signal Processing}, 92:\penalty0 2738--2746, 2012.

\bibitem[Maurer \& Pontil(2010)Maurer and Pontil]{DBLP:journals/tit/MaurerP10}
Maurer, Augusto and Pontil, Massimiliano.
\newblock K -dimensional coding schemes in hilbert spaces.
\newblock \emph{{IEEE} Trans. Information Theory}, 56\penalty0 (11):\penalty0
  5839--5846, 2010.

\bibitem[M\'emoli(2011)]{journals/focm/Memoli11}
M\'emoli, Facundo.
\newblock Gromov-wasserstein distances and the metric approach to object
  matching.
\newblock \emph{Foundations of Computational Mathematics}, 11\penalty0
  (4):\penalty0 417--487, 2011.

\bibitem[Mirkin(1996)]{opac-b1078049}
Mirkin, Boris~Grigorievitch.
\newblock \emph{Mathematical classification and clustering}.
\newblock Nonconvex optimization and its applications. Kluwer academic publ,
  Dordrecht, Boston, London, 1996.

\bibitem[Monge(1781)]{monge_81}
Monge, Gaspard.
\newblock M\'emoire sur la th\'eorie des d\'eblais et des remblais.
\newblock \emph{Histoire de l'Acad\'emie Royale des Sciences}, pp.\  666--704,
  1781.

\bibitem[Nadif \& Govaert(2008)Nadif and Govaert]{NADI08CI}
Nadif, M. and Govaert, G.
\newblock Algorithms for model-based block gaussian clustering.
\newblock In \emph{DMIN'08, the 2008 International Conference on Data Mining},
  2008.

\bibitem[Patrikainen \& Meila(2006)Patrikainen and Meila]{Patrikainen06}
Patrikainen, A. and Meila, M.
\newblock Comparing subspace clusterings.
\newblock \emph{IEEE Transactions on Knowledge and Data Engineering},
  18\penalty0 (7):\penalty0 902--916, July 2006.

\bibitem[Perrot et~al.(2016)Perrot, Courty, Flamary, and
  Habrard]{DBLP:conf/nips/PerrotCFH16}
Perrot, Micha{\"{e}}l, Courty, Nicolas, Flamary, R{\'{e}}mi, and Habrard,
  Amaury.
\newblock Mapping estimation for discrete optimal transport.
\newblock In \emph{NIPS}, pp.\  4197--4205, 2016.

\bibitem[Peyr{\'{e}} et~al.(2016)Peyr{\'{e}}, Cuturi, and
  Solomon]{DBLP:conf/icml/PeyreCS16}
Peyr{\'{e}}, Gabriel, Cuturi, Marco, and Solomon, Justin.
\newblock Gromov-wasserstein averaging of kernel and distance matrices.
\newblock In \emph{Proceedings of the 33nd International Conference on Machine
  Learning, {ICML} 2016, New York City, NY, USA, June 19-24, 2016}, pp.\
  2664--2672, 2016.

\bibitem[Rabin et~al.(2011)Rabin, Peyr\'e, Delon, and
  Bernot]{conf/scalespace/RabinPDB11}
Rabin, Julien, Peyr\'e, Gabriel, Delon, Julie, and Bernot, Marc.
\newblock Wasserstein barycenter and its application to texture mixing.
\newblock In \emph{Proceedings SSVM}, volume 6667, pp.\  435--446, 2011.

\bibitem[Rocci \& Vichi(2008)Rocci and Vichi]{rocci_08}
Rocci, R. and Vichi, M.
\newblock Two-mode multi-partitioning.
\newblock \emph{Computational Statistics and Data Analysis}, 52\penalty0
  (4):\penalty0 1984--2003, 2008.

\bibitem[Rubner et~al.(2000)Rubner, Tomasi, and
  Guibas]{Rubner:2000:EMD:365875.365881}
Rubner, Yossi, Tomasi, Carlo, and Guibas, Leonidas~J.
\newblock The earth mover's distance as a metric for image retrieval.
\newblock \emph{International Journal on Computer Vision}, 40\penalty0
  (2):\penalty0 99--121, 2000.

\bibitem[Schr\"odinger(1931)]{schrodinger}
Schr\"odinger, E.
\newblock Uber die umkehrung der naturgesetze.
\newblock \emph{Sitzungsberichte Preuss. Akad. Wiss. Berlin. Phys. Math.},
  144:\penalty0 144--153, 1931.

\bibitem[Shan \& Banerjee(2010)Shan and Banerjee]{shan_10}
Shan, Hanhuai and Banerjee, Arindam.
\newblock Residual bayesian co-clustering for matrix approximation.
\newblock In \emph{Proceedings of the {SIAM} International Conference on Data
  Mining, {SDM} 2010, April 29 - May 1, 2010, Columbus, Ohio, {USA}}, pp.\
  223--234, 2010.

\bibitem[Sinkhorn \& Knopp(1967)Sinkhorn and Knopp]{sinknopp_67}
Sinkhorn, Richard and Knopp, Paul.
\newblock Concerning nonnegative matrices and doubly stochastic matrices.
\newblock \emph{Pacific Journal of Mathematics}, 21:\penalty0 343--348, 1967.

\bibitem[Villani(2009)]{opac-b1129524}
Villani, C\'edric.
\newblock \emph{Optimal transport : old and new}.
\newblock Grundlehren der mathematischen Wissenschaften. Springer, Berlin,
  2009.

\bibitem[Wang et~al.(2009)Wang, Domeniconi, and Laskey]{pu2009latent}
Wang, Pu, Domeniconi, Carlotta, and Laskey, Kathryn.
\newblock Latent dirichlet bayesian co-clustering.
\newblock \emph{Machine Learning and Knowledge Discovery in Databases}, pp.\
  522--537, 2009.

\bibitem[Wilcoxon(1945)]{wilcoxon_45}
Wilcoxon, F.
\newblock {Individual Comparisons by Ranking Methods}.
\newblock \emph{Biometrics Bulletin}, 1\penalty0 (6):\penalty0 80--83, December
  1945.

\bibitem[Wyse \& Friel(2012)Wyse and Friel]{WyseF12a}
Wyse, Jason and Friel, Nial.
\newblock Block clustering with collapsed latent block models.
\newblock \emph{Statistics and Computing}, 22\penalty0 (2):\penalty0 415--428,
  2012.

\bibitem[Wyse et~al.(2014)Wyse, Friel, and Latouche]{wyse14}
Wyse, Jason, Friel, Nial, and Latouche, Pierre.
\newblock {Inferring structure in bipartite networks using the latent block
  model and exact ICL}.
\newblock \emph{ArXiv e-prints}, 2014.

\bibitem[Xu et~al.(2012)Xu, Bu, Chen, and Cai]{Xu:2012:EIC:2187836.2187840}
Xu, Bin, Bu, Jiajun, Chen, Chun, and Cai, Deng.
\newblock An exploration of improving collaborative recommender systems via
  user-item subgroups.
\newblock In \emph{Proceedings WWW}, pp.\  21--30, 2012.

\end{thebibliography}
\bibliographystyle{icml2017}

\end{document}